\documentclass[10pt,twocolumn,letterpaper]{article}

\usepackage{cvpr}
\usepackage{times}
\usepackage{epsfig}
\usepackage{graphicx}
\usepackage{amsmath}
\usepackage{amssymb}

\usepackage{subcaption}
\usepackage{color}
\usepackage{algorithm}
\usepackage{algpseudocode}
\usepackage{url}
\usepackage{multirow}
\usepackage{wrapfig}
\usepackage{xcolor}

\usepackage[pagebackref=true,breaklinks=true,letterpaper=true,colorlinks,bookmarks=false]{hyperref}

\cvprfinalcopy 

\def\cvprPaperID{1569} 

\algnewcommand\algorithmicforeach{\textbf{for each}}
\algdef{S}[FOR]{ForEach}[1]{\algorithmicforeach\ #1\ \algorithmicdo}

\newcommand{\yasu}[1]{}
\newcommand{\hang}[1]{}
\newcommand{\qi}[1]{}

\ifx\pdfoptionalwaysusepdfpagebox\relax\else
\fi

\ifcvprfinal\fi
\begin{document}

\title{RIDI: Robust IMU Double Integration}

\author{Hang Yan\\
Washington University in St. Louis\\
{\tt\small yanhang@wustl.edu}
\and
Qi Shan\\
Zillow Group\\
{\tt\small qis@zillow.com}
\and
Yasutaka Furukawa\\
Simon Fraser University\\
{\tt\small furukawa@sfu.ca}
}


\twocolumn[{%
\renewcommand\twocolumn[1][]{#1}%
\maketitle
\vspace{-1cm}
\begin{center}
    \centering
   \includegraphics[width=\textwidth]{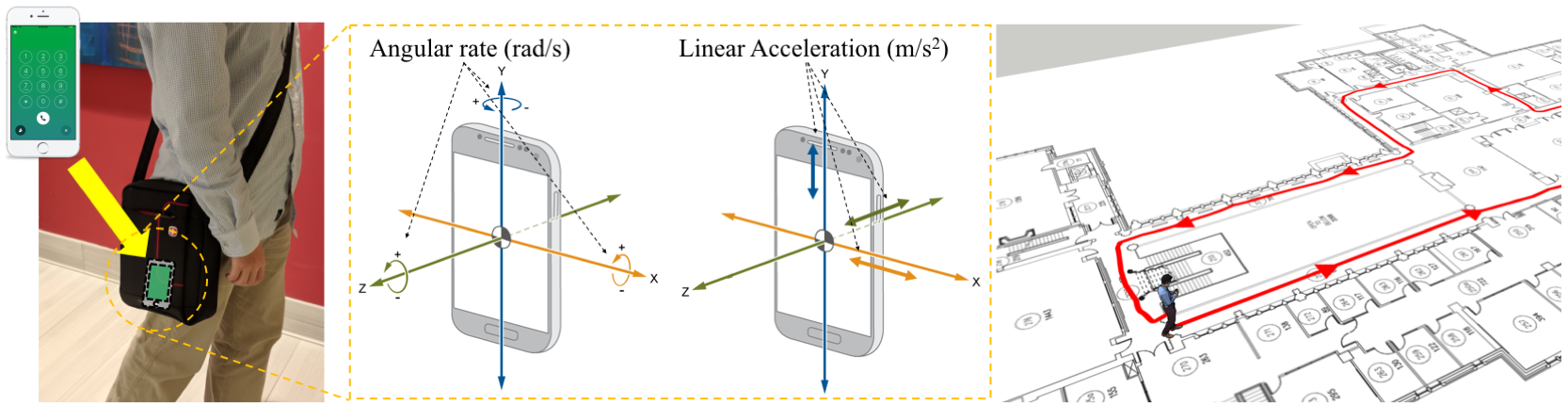}
 \captionof{figure}{Smartphones with motion sensors are ubiquitous in modern life. This paper proposes a novel data-driven approach for inertial navigation, which uses Inertial Measurement Unit (IMU) in every smartphone to estimate trajectories of natural human motions. IMU is energy-efficient and works anytime anywhere even for smartphones in your pockets or bags. 
 %
 }
\end{center}%
}]

\begin{abstract}
This paper proposes a novel data-driven approach for inertial navigation, which learns to estimate trajectories of natural human motions just from an inertial measurement unit (IMU) in every smartphone.
%
The key observation is that human motions are repetitive and consist of a few major modes (e.g., standing, walking, or turning). Our algorithm regresses a velocity vector from the history of linear accelerations and angular velocities, then corrects low-frequency bias in the linear accelerations, which are integrated twice to estimate positions.
%
We have acquired training data with ground-truth motions across multiple human subjects and multiple phone placements (e.g., in a bag or a hand).
The qualitatively and quantitatively evaluations have demonstrated that our algorithm has surprisingly shown comparable results to full Visual Inertial navigation.
%
To our knowledge, this paper is the first to integrate sophisticated machine learning techniques with inertial navigation, potentially opening up a new line of research in the domain of data-driven inertial navigation.
We will publicly share our code and data to facilitate further research.~\footnote{Project website: https://yanhangpublic.github.io/ridi/index.html}

\end{abstract}

\section{Introduction}

IMU double integration for motion estimation
has long been a dream for academic researchers and industry engineers. IMU is 1) energy-efficient, capable of running 24 hours a day without draining a battery;
and 2) works anywhere even inside a bag or a pocket.
%
The theory
is simple: given the device orientation (e.g., via Kalman filter on IMU signals), one subtracts the gravity from the device acceleration, integrates the residual accelerations once to get velocities, and integrates once more to get positions.
%
Unfortunately, small sensor errors or biases explode quickly in the double integration process. Such systems do not work in practice, unless one uses a million dollar military-grade IMU unit, often found on submarines.
Our key idea is that human motions are repetitive and consist of a small number of major modes. Pedometry precisely exploits this property to estimate the travel distance by step-counting. Together with standard IMU-based rotation estimation, it would appear relatively easy to estimate reasonable walking trajectories just from IMUs. However, this approach assumes that the device rotation is exactly aligned with the walking direction, which is not the case for our smartphones in our hands, bags, or leg pockets. The approach also fails for side motions or backward motions.

Our algorithm, dubbed Robust IMU Double Integration (RIDI), takes a data driven approach and learns to regress the instantaneous motion (i.e., a velocity) from IMU signals, while automatically adjusting 
arbitrary device rotations with respect to a body. More precisely, RIDI
regresses a velocity vector from the history of linear accelerations and angular velocities, then corrects low-frequency errors in the linear accelerations to be compatible with the regressed velocities.
A standard double integration is used to estimate the trajectory from the corrected linear accelerations.
%
%

We have acquired IMU sensor data across six human subjects with four popular smartphone placements. The ground-truth trajectories are obtained by a Visual Inertial Odometry system (i.e., a Google Tango phone, Lenovo Phab2 Pro)~\cite{tango}. Our datasets consist of various motion trajectories over 100 minutes at 200Hz. Our experiments have shown that RIDI produces motion trajectories comparable to the ground truth, with mean positional errors below 3\%. 


To our knowledge, this paper is the first to integrate sophisticated machine learning techniques with inertial navigation, and could start a new line of research in data-driven Inertial Navigation.
Commercial implications of the proposed research are also significant.
IMUs are everywhere on the market, inside smartphones, tablets, or emerging wearable devices (e.g., Fitbit or Apple Watch). Almost everybody always carries one of these devices, for which RIDI could provide precise motion information with minimal additional energy consumption, a potential to enable novel location-aware services in broader domains.
We will publicly share out code and data to facilitate further research.
%

%



\section{Related Work}

Visual SLAM (V-SLAM)~\cite{recent_slam_survey} has made remarkable progress in the last decade~\cite{ptam,orb_slam,dtam,dso}, enabling a robust real-time system for indoors or outdoors up to a scale ambiguity.
%
Visual-inertial SLAM (VI-SLAM) combines V-SLAM and IMU, resolving the scale ambiguity and making the system further robust. VI-SLAM is used in many successful products
such as 
Google Project Tango~\cite{tango}, Google ARCore~\cite{arcore}, Apple ARKit~\cite{arkit} or Microsoft Hololens~\cite{hololens}. While being successful, the system suffers from two major drawbacks: 1) a camera must have a clear light-of-sight under well-lit environments all the time, and 2) the recording and processing of video data quickly drain a battery.
IMU-based rotation estimation has been successful, used in 
many recent Virtual Reality applications such as Google Cardboard VR~\cite{google_cardboard} or Samsung Gear VR~\cite{samsung_gear_vr}.
IMU has also been proven effective for gait recognition~\cite{mansur2014gait}, activity recognition~\cite{von2017sparse}, or step-counting~\cite{step_counting,kasebzadeh2016improved}.
Step-counting, in particular, can lead to travel distance estimation, but the motion estimation is a challenge as the device orientation and the motion direction are not always aligned.
%
In general, IMU-based position estimation 
still requires impractical assumptions, such as IMU on a foot~\cite{yun2007self} or presence of map data~\cite{li2012reliable}.
%
%
%
This paper integrates machine learning and inertial navigation to propose a robust data-driven approach without heuristics or impractical assumptions.

WiFi-based position tracking could also be a low-energy anytime-anywhere solution~\cite{lim2007real,bahl2000radar,WifiSlamOld,WifiSlam}. However, these technologies are orthogonal to our work and we do not consider them as competing methods. State-of-the-art WiFi-based tracking system, WiFiSlam~\cite{WifiSlam}, critically depends on inertial navigation, and this paper directly benefits WiFi-based tracking systems. 
%


%


\section{Inertial 3D Motion Database}

\begin{figure}[tb]
\centering \includegraphics[width=0.49\textwidth]{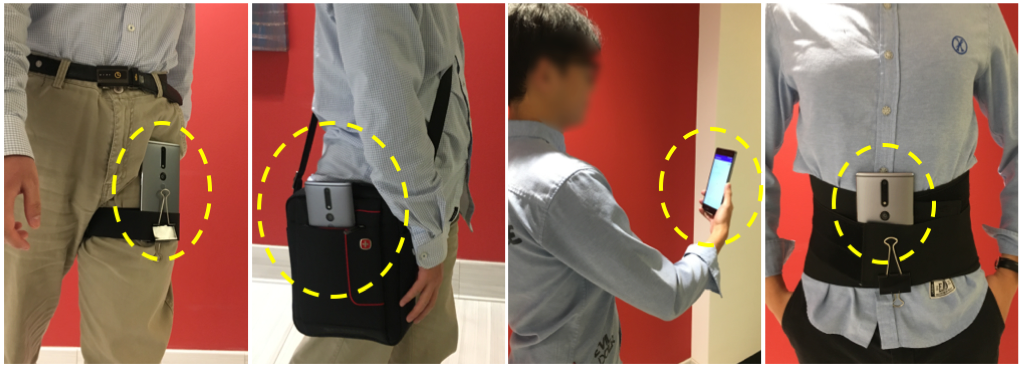}
\caption{We place a Tango phone in four popular configurations to collect training data. The ground-truth motions come from Visual Inertial Odometry, and we have carefully designed the placements to make the camera always visible.
From left to right: 1) in a leg pocket, 2) in a bag, 3) in a hand, or
4) on a body (e.g., for officers).
}
\label{fig:frames}
\end{figure}

One contribution of the paper is a database of IMU sensor measurements and 3D motion trajectories across multiple human subjects and multiple device placements. We have used a Google Tango phone, Lenovo Phab2 Pro, to record linear accelerations, angular velocities, gravity directions, device orientations (via Android APIs), and 3D camera poses.
%
The camera poses come from the Visual Inertial Odometry system on Tango, and we make sure that the camera has a clear field-of-view all the time (See Fig.~\ref{fig:frames}).


We have collected more than 100 minutes of data at 200Hz from six human subjects under four popular smartphone placements with various motion types including walking forward/backward, side motion, or acceleration/deceleration. Asynchronous signals from various sources are synchronized into the time-stamps of Tango poses via linear interpolation as a pre-processing step.


\section{Algorithm}

The proposed algorithm, dubbed Robust IMU Double Integration (RIDI), consists of two steps.
First, RIDI regresses a velocity vector from angular velocities and linear accelerations (i.e., accelerometer readings minus gravity). Second, RIDI 
estimates low-frequency corrections in the linear accelerations so that their integrated velocities match the regressed values. Corrected linear accelerations are integrated to estimate positions.
We assume that subjects walk on a flat floor. The regression and the position estimation are conducted on a 2D horizontal plane. We now explain a few coordinate frames and the details of the two steps.


\subsection{Coordinate frames}
\setlength{\columnsep}{6pt}%
\setlength{\intextsep}{0pt}
\begin{wrapfigure}{r}{0.55\columnwidth}
\centerline{
    \includegraphics[width=0.55\columnwidth]{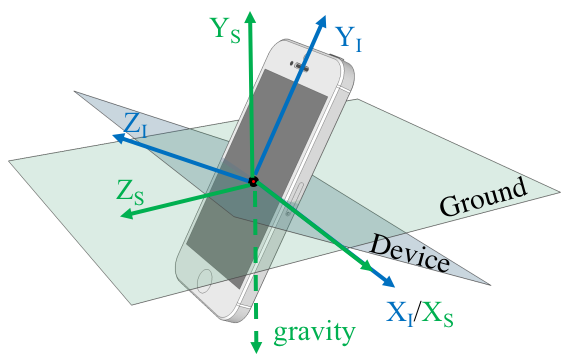}
}
\end{wrapfigure}

We consider three coordinate frames in our algorithm. The first one is the world coordinate frame $W$, in which the output positions are estimated. $W$ is set to be the global coordinate frame from the Android API at the first frame.
The second one is the IMU/device coordinate frame $I$ (marked with blue arrows in the right figure) in which IMU readings are provided by the system APIs.
Lastly, we leverage the gravity direction from the system
to define our stabilized-IMU frame $S$, where the device pitch and roll are eliminated from $I$ by aligning its y-axis with the gravity vector (see the green arrows in the right figure). This coordinate frame makes our regression task easier, since the regression becomes independent of the device tilting and rolling.

\subsection{Learning to regress velocities}
\begin{figure}[tb]
\includegraphics[width=0.48\textwidth]{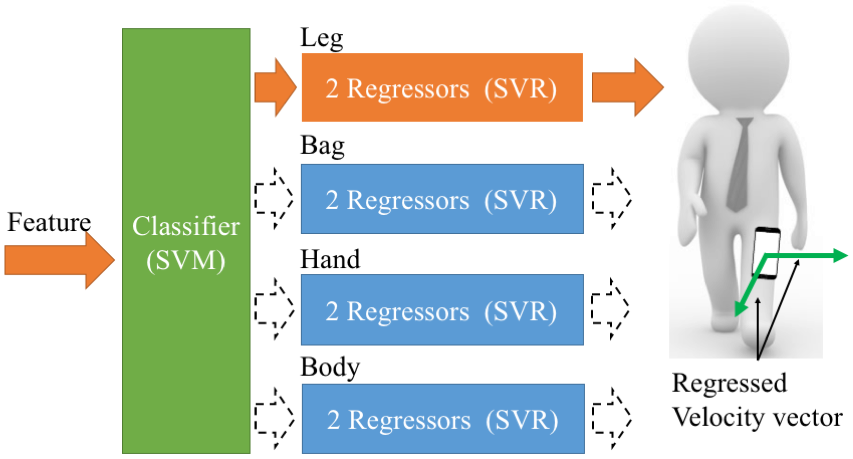}
\caption{Our cascaded regression consists of one SVM and eight SVRs. SVM classifies the phone placement from the four types. Two type-specific SVRs predict a 2D velocity in the stabilized-IMU frame, ignoring the vertical direction.
}
\label{fig:cascade_model}
\end{figure}

\begin{figure*}[t]
\includegraphics[width=\textwidth]{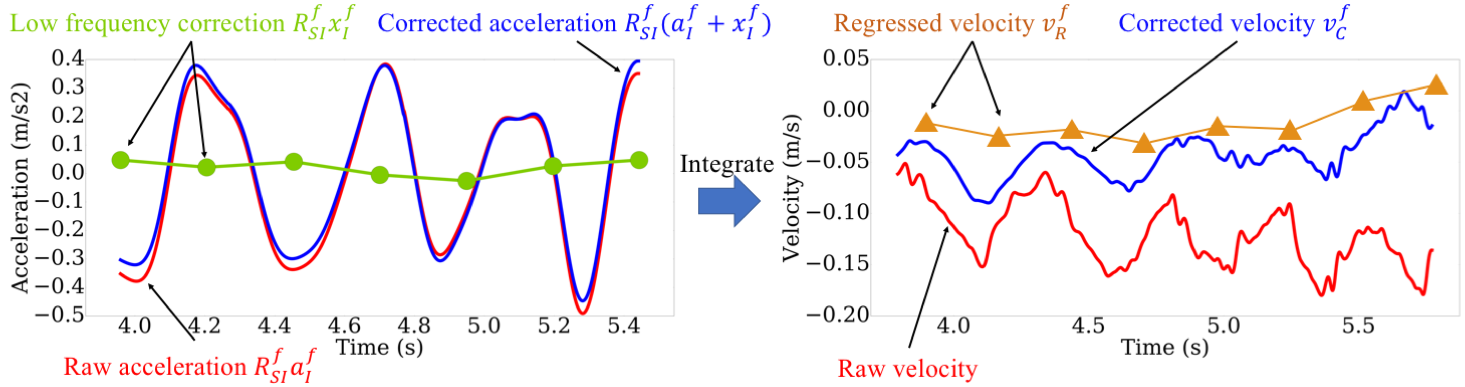}
\caption{
Robust IMU double integration process. Our approach directly models the errors (green on the left) in the linear acceleration as a piecewise linear (thus low-frequency) function. We estimate parameters of this correction function so that the integration of the corrected linear accelerations (blue on the right) matches the regressed velocities (brown on the right). 
}
\label{fig:model}
\end{figure*}

We learn to regress velocities in the stabilized IMU frame $S$.
For each training sequence, we transform device poses (in $W$), and angular velocities and linear accelerations (in $I$) into $S$. 
The central difference generates velocity vectors from the transformed device poses (ignoring the vertical direction).
To suppress high-frequency noise, we apply Gaussian smoothing with $\sigma=2.0$ frames to 6 IMU channels, and with $\sigma=30.0$ frames to 2 velocity channels, respectively. We concatenate smoothed angular velocities and linear accelerations from the past 200 frames (i.e., 1 second) to construct a $1200$ dimensional feature vector.

People carry smartphones in different ways, exhibiting different IMU signal patterns.
We assume that a phone is either 1) in a leg pocket, 2) in a bag, 3) hand-held, or 4) on body, and exploit this knowledge to propose a cascaded regression model (See Fig.~\ref{fig:cascade_model}).
%
%
\begin{table}
\caption{Hyper-parameters for SVRs are found by the grid search: (1) $C$ within a range of $[0.1, 100.0]$ with a multiplicative increment of 10; and (2) $\epsilon$ within a range of $[0.001, 1.0]$ with a multiplicative increment of 10.}
\centering
\begin{tabular}{|c||c|c|c|c|}
\hline
& Leg & Bag & Hand & Body \\
\hline \hline
$C$ & 1.0 & 10.0 & 10.0 & 1.0 \\
\hline
$\epsilon$  & 0.001 & 0.01 & 0.001 & 0.001 \\
\hline
\end{tabular}

\label{table:grid_search}
\end{table}
More precisely, a Support Vector Machine (SVM) first classifies the placement to be one of the above four types, then two type-specific $\epsilon$-insensitive Support Vector Regression (SVR)~\cite{smola2004tutorial} models estimate two velocity values independently (ignoring the vertical direction). 
The hyper-parameters for each of the SVM/SVR models are tuned independently by the grid search and 3-fold cross validation,
based on the mean squared error on the regressed velocities. The grid-search finds the soft-margin parameter $C=10.0$ for SVM. Table~\ref{table:grid_search} summarizes the chosen parameters for SVR models.

\subsection{Correcting acceleration errors}

Predicted velocities provide effective cues in removing sensor noises and biases.\footnote{Direct integration of the predicted velocities would produce positions but perform worse (See Sect.~\ref{sect:results} for comparisons).}
%
The errors come from various sources (e.g., IMU readings, system gravities, or system rotations) and interact in a complex way. 
We make a simplified assumption and model all the errors as a low-frequency bias in the linear acceleration.
This approach is not physically grounded, but bypasses explicit noise/bias modeling and turns our problem into simple linear least squares.

We model the bias in the linear acceleration in the IMU/device coordinate frame $I$. 
To enforce the low-frequency characteristics, we represent the bias as linear interpolation of correction terms $x^f_I$ at sub-sampled frames ($\mathcal{F}_1$), in particular, one term every 50 frames~\cite{zhou2014simultaneous}. With abuse of notation, we also use $x^f_I$ to denote interpolated acceleration correction (e.g., $x^{11}_I = 0.8 x^{1}_I + 0.2 x^{51}_I$).

Our goal is to estimate $\{x^f_I\}$ at $\mathcal{F}_1$ by minimizing the discrepancy between the corrected velocities $(v^f_C)$ and the regressed velocities $(v^f_R)$ at sub-sampled frames $\mathcal{F}_2$ (once every 50 frames, to avoid evaluating SVRs at every frame for efficiency). The discrepancy is measured in the stabilized IMU frame $S$.
%
\begin{equation}
\begin{split}
\min_{\{x^1_I, x^{51}_I, \cdots\}} 
\sum_{f\in \mathcal{F}_2} \left\| v^f_C - v^f_R\right\|^2 + \lambda  \sum_{f\in \mathcal{F}_1}  \left\| x^f_I \right\|^2, \label{eq:optimization} \\
v^f_C = \mathcal{R}^f_{SW}\sum_{f^\prime=1}^f \mathcal{R}^{f^\prime}_{WI}\left( a^{f^\prime}_{I} + x^{f^\prime}_I \right).
\end{split}
\end{equation}
$a^f_{I}$ denotes the raw linear acceleration in $I$. $\mathcal{R}_{AB}$ denotes the rotation that transforms a vector in coordinate frame $\mathcal{B}$ to $\mathcal{A}$.
The corrected velocity $(v^f_C)$ is simply the integration of the corrected accelerations, except that the summation must occur in a consistent coordinate frame, for example $W$ (but not $S$, which changes every frame).~\footnote{We assume zero-velocity at the first frame, which is the case for our datasets. Relaxing this assumption is our future work.}
%
%
%

The first term minimizes the velocity discrepancy.
Note that our regressor estimates 
the horizontal velocity, namely only the two entries in $v^f_R$ except the vertical direction. We assume that subjects walk on the flat surface, and hence, fix the vertical component of $v^f_R$ to be 0.
%
%
%
The second term enforces $l_2$ regularization, which allows us to balance the velocity regression and the raw IMU signals.
When $\lambda$ is 0, the system simply integrates the regressed velocities without using raw IMU data. When $\lambda$ is infinity, the system ignores the regressed velocities and performs the naive IMU double integration. 
We have used $\lambda=0.1$ in our experiments. Double integration of the corrected accelerations produces our position estimations.

\section{Implementation}


We have implemented the proposed system in C++ with third party libraries including OpenCV~\cite{opencv}, Eigen~\cite{eigen}, and Ceres Solver~\cite{ceres-solver}. Note that our optimization problem (\ref{eq:optimization}) has a closed form solution, but we use Ceres for the ease of implementation. We have used a desktop PC equipped with a Intel I7-4790 CPU and 32GB RAM.



We have presented the algorithm as an offline batch method for clarity. It is fairly straightforward to implement an online algorithm, which has been used in all our experiments.
The online algorithm keeps track of two threads. The first thread updates correction terms once every 200 frames using the latest 1,000 frames.
The second thread conducts double-integration every frame to produce a position estimation. The correction terms are initialized to 0 before being updated by the first thread. 
The two expensive steps are the SVR regression and the correction optimization, which takes 19ms and 15ms on the average, respectively.
Our system processes 10,000 frames within 5 seconds, effectively achieving 2,000 fps on a desktop PC.


\begin{figure*}[tb]
\centering \includegraphics[width=0.9\textwidth]{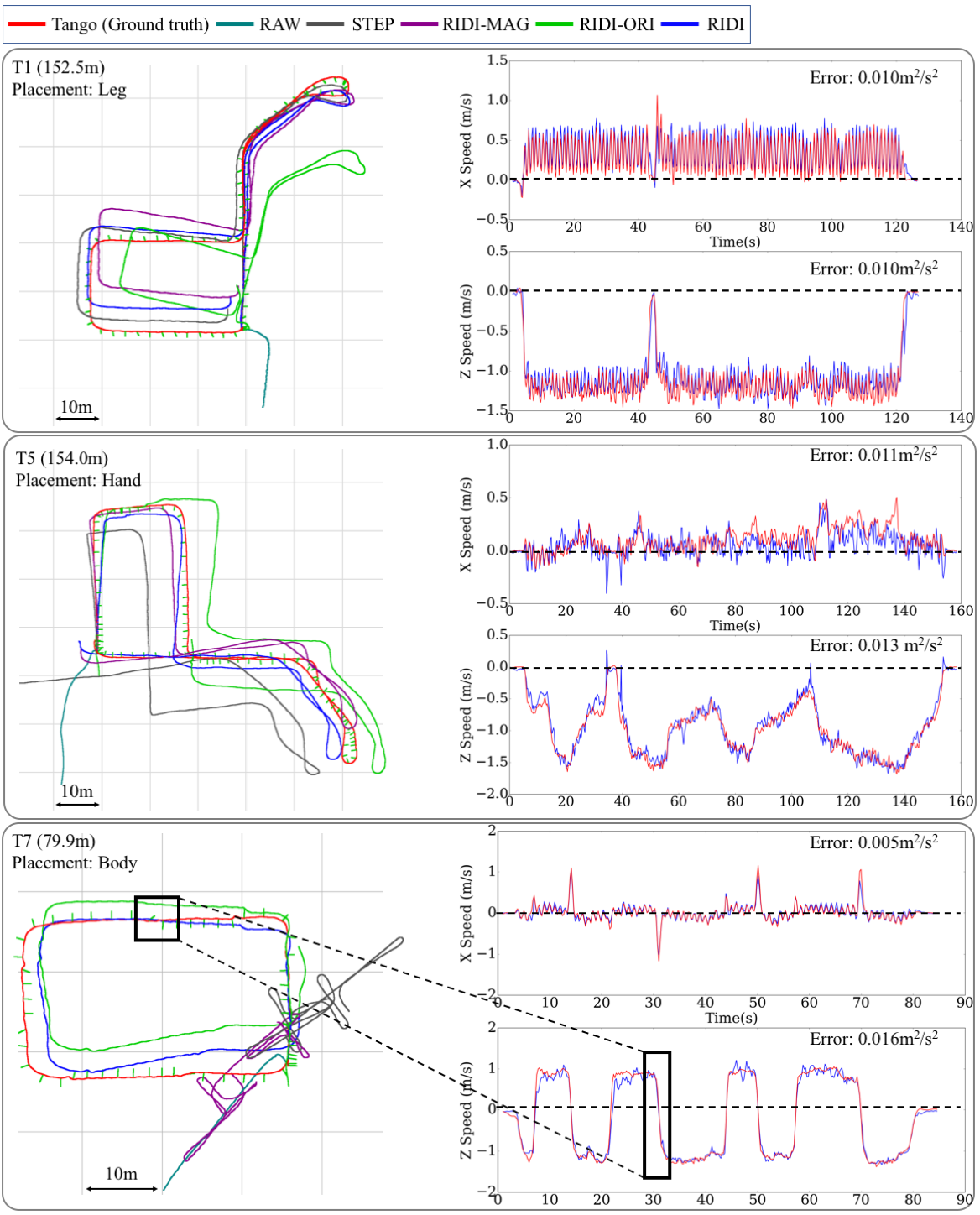}
\caption{
Left: Motion trajectories from Tango,  competing methods, and RIDI. Right: Regressed velocity vectors and their mean squared errors (MSE). Short green segments indicate headings of device's X axis. The naive double integration (RAW) fails in many examples. In T5 (middle row), the subject frequently changes the speed, where STEP and RIDI-ORI  produce large errors for inaccurate speed magnitudes. 
In T7 (bottom row), the subject mixes different walking patterns including 4 backward motions (the black rectangle is one place), where STEP and RIDI-MAG fails for not inferring velocity directions.
}
\label{fig:tracking}
\end{figure*}

\begin{table*}[tb]
\caption{
Positional accuracy evaluations. Each entry shows the mean positional error 
and its ratio (inside a parentheses) with respect to the trajectory distance.
The blue and the brown numbers show the best and the second best results.
}
\label{tab:indoor}
\centering
\begin{tabular}{|c|c|c|c|c|c|c|}
\hline 
Sequence & Placement & RAW [m] & STEP [m] & RIDI-MAG [m] & RIDI-ORI [m] & RIDI [m] \\
\hline
T1 & Leg & 55.92(36.95\%) & \color{brown}1.64(1.08\%) & 2.37(1.57\%) & 6.22(4.11\%) & \color{blue}1.50(0.99\%)\\
\hline
T2 & Leg & 57.06(83.68\%) & \color{blue}0.76(1.11\%) & 1.31(1.92\%) & 3.29(4.82\%)  & \color{brown}1.23(1.81\%)\\
\hline
T3 & Bag & 71.93(46.43\%) & 6.70(4.33\%) & \color{brown}4.44(2.87\%) & 6.40(4.13\%) & \color{blue}4.07(2.63\%)\\
\hline
T4 & Bag & 18.18(23.98\%) & 3.13(4.13\%) & \color{brown}0.80(1.06\%) & 3.14(4.15\%) & \color{blue}0.60(0.79\%)\\
\hline
T5 & Hand & 180.03(116.91\%) & 9.32(6.05\%) & \color{brown}2.54(1.65\%) & 8.37(5.44\%) & \color{blue}0.91(0.59\%)\\
\hline
T6 & Hand & 28.93(61.09\%) & 5.56(11.74\%) & 4.23(8.94\%) & \color{brown}3.33(7.02\%) & \color{blue}1.06(2.23\%)\\
\hline
T7 & Body & 56.06(70.17\%) & 16.28(20.37\%) & 10.87(13.60\%) & \color{brown}2.88(3.61\%) & \color{blue}1.77(2.22\%)\\
\hline
T8 & Body & 20.14(29.41\%) & 0.96(1.41\%) & \color{brown}0.75(1.09\%) & 2.02(2.95\%) & \color{blue}0.77(1.12\%)\\
\hline
\end{tabular}
\end{table*}

\begin{table}
\caption{The average MPE (as a ratio against the trajectory distance) over the testing sequences with different $\lambda$.
}
\label{tab:weight}\centering
\begin{tabular}{|c|c|c|c|c|c|}
\hline
$\lambda$ & 0.0001 & 0.001 & 0.1 & 1.0 & 10,000\\
\hline
MPE  & 11.62\% & 1.49\% & 1.45\% & 1.47\% & 33.98\%\\
\hline
\end{tabular}

\end{table}

\begin{figure*}
  \centering
  \includegraphics[width=\textwidth]{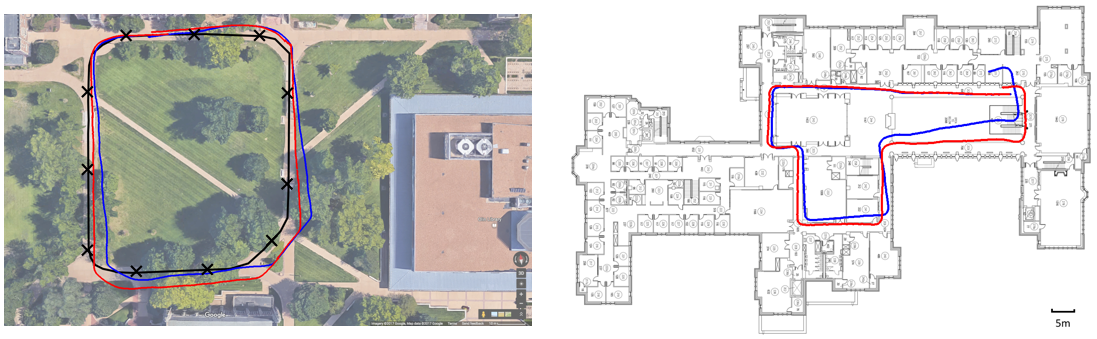}
  \caption{Overlaying the trajectories with the online digital map (from Google Maps) or  the floorplan image with the estimated scale. The red line marks the trajectory given by the Tango system and the blue line marks the trajectory given by our system. The accuracy of the Tango system degrades at outdoors in our experiments, so we manually drew the actual walking path with the black line at the left.}
  \label{fig:map_overlay}
\end{figure*}

\begin{figure*}[t]
\centering
\includegraphics[width=\textwidth]{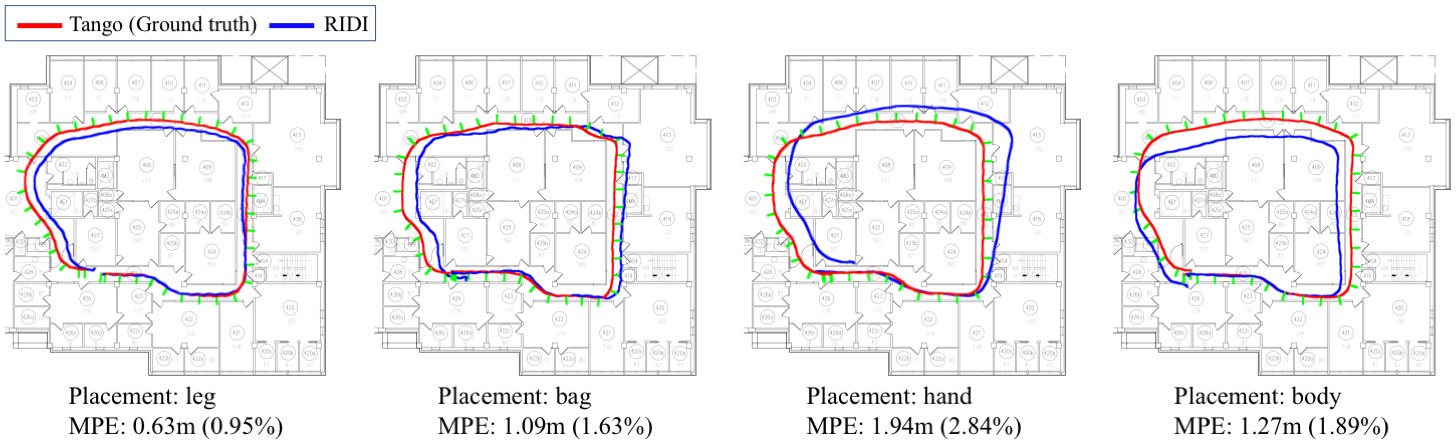}
\caption{
Generalization to an unseen device (Google Pixel XL). 
}
\label{fig:pixel}
\end{figure*}

\begin{figure*}[!ht]
\centering
\includegraphics[width=0.9\textwidth]{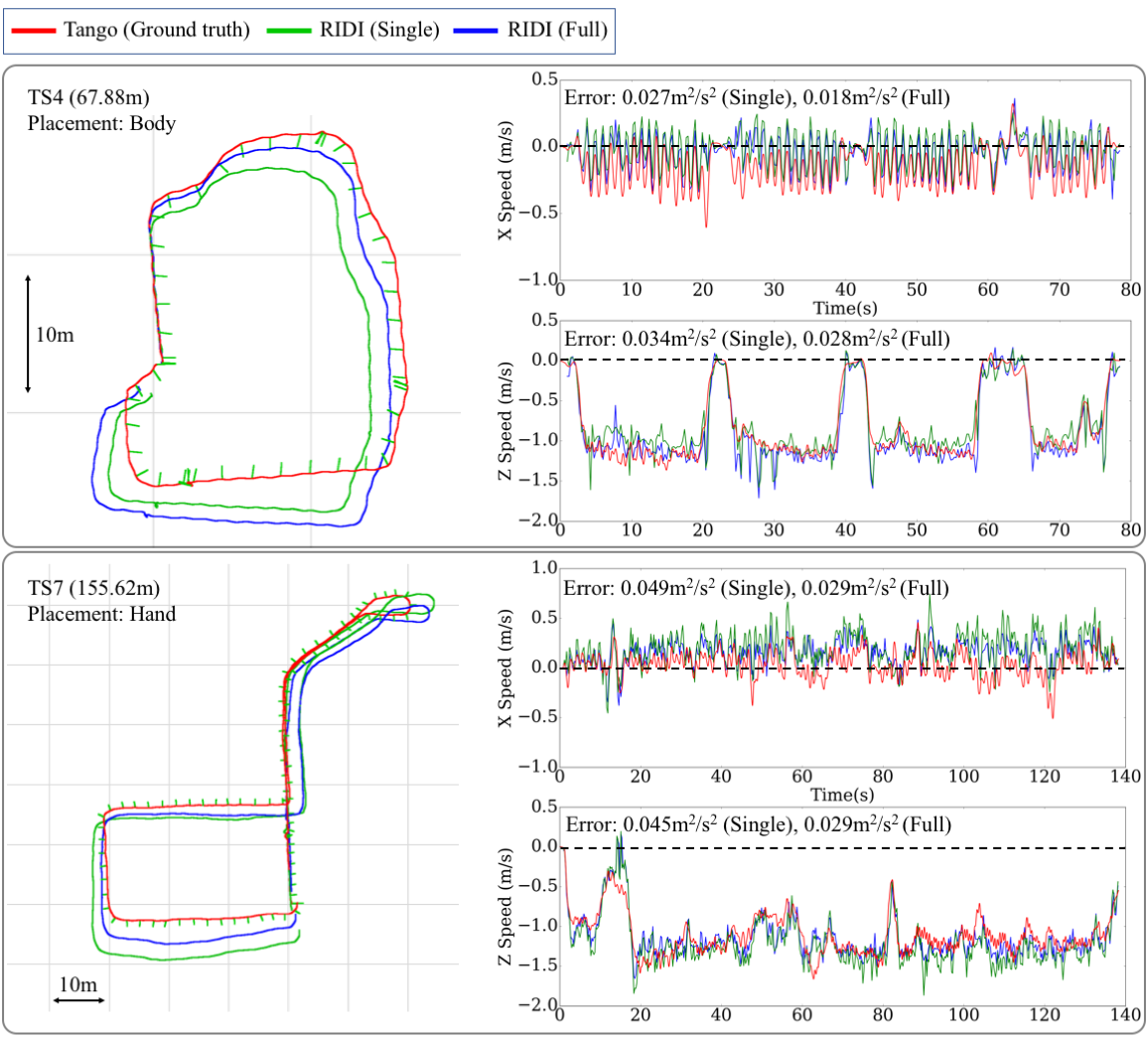}
\caption{
Generalization to unseen subjects. We varied the number of human subjects in the training data and evaluated two RIDI models for unseen testing subjects. RIDI (single) uses training data only from one subject, while RIDI (Full) uses training data from the four subjects.
}
\label{fig:cross_subject}
\end{figure*}

\begin{table*}[!ht]
\caption{
Generalization to unseen subjects. The forth and fifth columns are the mean squared errors on the regressed velocities along the two horizontal axes. Last two columns are the mean positional errors (MPE) in meter. The model trained on more subjects generalizes better.}
\label{tab:cross_subject}
\centering
\begin{tabular}{|c|c|c||c|c||c|c|}
\hline
Sequence & Subject & Placement & Reg. Error (Single) & Reg. Error (Full) & MPE (Single) & MPE (Full)\\
\hline
TS1 & S5 & Leg & (0.051, 0.318) & \color{blue}(0.027, 0.089) & 5.98(8.25\%) & \color{blue}3.51(4.83\%)\\
\hline
TS2 & S5 & Bag & (0.108, 0.101) & \color{blue}(0.062, 0.060) & 3.81(5.03\%) & \color{blue}2.67(3.52\%)\\
\hline
TS3 & S5 & Hand & (0.007, 0.025) & \color{blue}(0.005, 0.017) & 1.50(2.22\%) & \color{blue}1.27(1.89\%)\\
\hline
TS4 & S5 & Body & (0.027, 0.034) & \color{blue}(0.018, 0.028) & 2.20(3.24\%) & \color{blue}2.16(3.18\%)\\
\hline
TS5 & S6 & Leg & (0.056, 0.308) & \color{blue}(0.050, 0.110) & 6.00(8.69\%) & \color{blue}3.28(4.75\%)\\
\hline
TS6 & S6 & Bag & (0.316, 0.427) & \color{blue}(0.195, 0.322) & 11.79(11.36\%) & \color{blue}8.36(8.05\%)\\
\hline
TS7 & S6 & Hand & (0.049, 0.045) & \color{blue}(0.029, 0.029) & 2.41(1.55\%) & \color{blue}2.18(1.40\%)\\
\hline
TS8 & S6 & Body & (0.024, 0.126) & \color{blue}(0.019, 0.077) & 3.90(7.73\%) & \color{blue}1.97(3.911\%)\\
\hline
\end{tabular}
\end{table*}
\section{Experimental results} \label{sect:results}\label{sec:tracking}
We have acquired 60 motion sequences over 4 human subjects (marked as S1-S4), 4 different phone placements, and a variety of motion types.
We have randomly selected 40 sequences for training and the remaining 20 sequences for testing. We have created one data sample per 10 frames, resulting in 67,649 training samples and 44,236 testing samples. 
We have also acquired additional sequences from two unseen human subjects and one unseen device for testing.



\subsection{Position evaluations}

\noindent \textbf{Baseline comparisons:}
Table~\ref{tab:indoor} summarizes the quantitative evaluations on the accuracy of the final positions over 8 testing sequences (marked as T1-T8). We compared our method against 4 competing methods:

\vspace{0.2cm}
\noindent $\bullet$ \textbf{RAW} denotes the naive double integration with the raw linear accelerations.

\noindent $\bullet$ \textbf{STEP} denotes the standard step-counting method. Step counts are provided by the Android API.
We use the ground-truth motion to compute the length of each step, and use their average to determine the step-length to be used by this method for each sequence.
The motion direction is set to the device rotation given by the Android API.~\footnote{We disable the magnetometer in our experiments, which damages rotation estimations due to instable magnetic fields.}


\noindent $\bullet$ \textbf{RIDI-MAG} is a variant of the proposed method. The regressed velocity vector consists of  the magnitude and direction information. 
RIDI-MAG keeps the velocity magnitude, while replacing its direction by the system rotation through the Android API. RIDI-MAG cannot compensate for the device rotations with respect to the body.


\noindent $\bullet$  \textbf{RIDI-ORI} is the same as RIDI-MAG except that it keeps the regressed direction, while replacing the regressed magnitude by the average of the ground-truth values for each sequence.

\vspace{0.2cm}
\noindent
For all the experiments, we align each motion trajectory to the ground-truth by computing a 2D rigid transformation that minimizes the sum of squared distances for the first 10 seconds (2,000 frames).
Table~\ref{tab:indoor} illustrates that RIDI outperforms all the other baselines in most sequences, and achieves mean positional errors (MPE) less than 3.0\% of the total travel distance, that is, a few meters after 150 meters of walking. 
Figure~\ref{fig:tracking} illustrates a few representative examples with regressed velocities. T5 is a case, in which the subject frequently changes the walking speeds. RIDI-ORI fails for assuming a constant speed and STEP fails for inaccurate stride lengths. T6 and T7 are cases, in which the subjects mix different walking patterns, including backward motions. Only RIDI and RIDI-ORI, which infer motion directions, perform well. 
Please refer to the supplementary material for more results and visualizations.

\vspace{0.1cm} \noindent \textbf{Scale consistency:}
One of the key advantages of the inertial or visual-inertial navigation is that the reconstruction is up to a metric-scale, which is not the case for image-only techniques such as visual-SLAM. Figure~\ref{fig:map_overlay} shows that our trajectories are well aligned over a satellite or a floorplan image. We adjusted the scales (meters per pixel) based on the rulers, and manually specified
the starting point and the initial orientation.

\vspace{0.1cm} \noindent \textbf{Parameter $\lambda$:} Table~\ref{tab:weight} shows the impact of the parameter $\lambda$ in Equation~\ref{eq:optimization}, suggesting that it is important to integrate the velocity regression with the raw IMU acceleration data. Neither the regressed velocities (small $\lambda$) nor the naive double integration (large $\lambda$) performs well on its own. From this result, we set $\lambda=0.1$ as our default parameter value.



\subsection{Velocity evaluations}
Our cascaded velocity regression achieves the mean squared errors of $0.016$ [m$^2$/s$^2$] and $0.015$ [m$^2$/s$^2$] on the X and Z axes on the testing set, respectively.
%
We have also calculated the accuracy of the SVM classifier on the placement types, where the training and the testing accuracies are $94.70\%$ and $93.65\%$, respectively.
Lastly, we have evaluated the all-in-one regression model by SVR without the placement classification. The mean squared errors on the X and Z axes are $0.32$ [m$^2$/s$^2$] and $0.54$ [m$^2$/s$^2$], respectively, which are much worse than our cascaded model. Acquiring more training data and evaluating the accuracy of more data-hungry methods such as deep neural networks is one of our future works.


\subsection{Generalization}
\noindent \textbf{Unseen devices:} Considering the impact to commercial applications, the generalization capability to unseen devices is of great importance. We have used another device (Google Pixel XL) to acquire additional testing sequences, while the subjects also carried the Tango phone to obtain ground truth trajectories. The sequence contains a quick rotation motion at the beginning to generate distinctive peaks in the gyroscope signals, which are used to synchronize  data from the two devices.
We register the estimated trajectories to the ground-truth by the same process as before.
Figure~\ref{fig:pixel} shows that our system generalizes reasonably well under all placement types, in particular, still keeping the mean positional errors below 3\%.



\vspace{0.1cm} \noindent \textbf{Unseen subjects:} The last experiment evaluates the generalization capability to unseen subjects (marked as S5 and S6).
These two subjects have no prior knowledge of our project and we asked them to walk in their own ways.
%
We have trained two RIDI models with different training sets. RIDI (Single) is trained on data only from one subject (S1). RIDI (Full) is trained on data from the four subjects (S1-S4). For fair comparisons, we have down-sampled both training sets to 22,000 samples.
%
Figure~\ref{fig:cross_subject} and Table~\ref{tab:cross_subject} demonstrate that the Full model generalizes well, in particular, below 5\% MPE in most cases. However, the system performs worse in some sequences, for example, 8 meter positional errors after 100 meters of walking, which is too large for many indoor applications.
Another important future work is to push the limit of the generalization capability by collecting more data and designing better regression machineries.


\section{Conclusion}
The paper proposes a novel data-driven approach for inertial navigation that robustly integrates linear accelerations to estimate motions. Our approach exploits patterns in natural human motions, learns to regress a velocity vector, then corrects linear accelerations 
via simple linear least squares, which are integrated twice to estimate positions.
%
Our IMU-only navigation system is energy efficient and works anywhere even inside a bag or a pocket, yet achieving 
comparable accuracy to a full visual inertial navigation system to our surprise.
%
%
Our future work is to collect a lot more training data across more human subjects on more devices, and learn a universal velocity regressor that works for anybody on any device. Another important future work is to deploy the system on computationally less powerful mobile devices.
The impact of the paper to both scientific and industrial communities could be profound. This paper has a potential to open up a new line of learning based inertial navigation research. Robust anytime-anywhere navigation system could immediately benefit a wide range of industrial applications through location-aware services including online advertisements, digital mapping, navigation, and more.

\section{Acknowledgement}
This research is partially supported by National Sci-
ence Foundation under grant IIS 1540012 and IIS 1618685,
Google Faculty Research Award.


{\small
\bibliographystyle{ieee}
\bibliography{egbib}
}

\end{document}